\documentclass{article}
\usepackage{ijcai22}

\usepackage{times}
\usepackage{soul}
\usepackage{url}
\usepackage[hidelinks]{hyperref}
\usepackage[utf8]{inputenc}
\usepackage[small]{caption}
\usepackage{graphicx}
\usepackage{amsmath}
\usepackage{amsthm}
\usepackage{booktabs}
\usepackage{algorithm}
\usepackage{algorithmic}
\usepackage{comment}
\usepackage{multirow}

\title{On Knowledge Editing in Federated Learning: Perspectives, \\Challenges, and Future Directions}

\author{
Leijie Wu$^1$
\and
Song Guo$^{1,2}$\and
Junxiao Wang$^1$\and
Zicong Hong$^1$\and
Jie Zhang$^1$\And
Jingren Zhou$^3$
\affiliations
$^1$The Hong Kong Polytechnic University\\
$^2$The Hong Kong Polytechnic University Shenzhen Research Institute\\
$^3$Alibaba Group
\emails
lei-jie.wu@connect.polyu.hk,
\{song.guo, junxiao.wang\}@polyu.edu.hk,
zicong.hong@connect.polyu.hk,
jie-comp.zhang@polyu.edu.hk,
jingren.zhou@alibaba-inc.com
}
\begin{document}

\maketitle

\begin{abstract}
\label{Abstract}
%junxiao-2023/1/21
As Federated Learning (FL) has gained increasing attention, it has become widely acknowledged that straightforwardly applying stochastic gradient descent (SGD) on the overall framework when learning over a sequence of tasks results in the phenomenon known as ``catastrophic forgetting''.
Consequently, much FL research has centered on devising federated increasing learning methods to alleviate forgetting while augmenting knowledge.
On the other hand, forgetting is not always detrimental.
The selective amnesia, also known as federated unlearning, which entails the elimination of specific knowledge, can address privacy concerns and create additional ``space'' for acquiring new knowledge.
%
%The selective amnesia namely federated unlearning -- the process of removing specific knowledge -- can address privacy concerns even leave more ``space'' for learning new knowledge.
%
However, there is a scarcity of extensive surveys that encompass recent advancements and provide a thorough examination of this issue. 
In this manuscript, we present an extensive survey on the topic of knowledge editing (augmentation/removal) in Federated Learning, with the goal of summarizing the state-of-the-art research and expanding the perspective for various domains.
Initially, we introduce an integrated paradigm, referred to as Federated Editable Learning (FEL), by reevaluating the entire lifecycle of FL. 
Secondly, we provide a comprehensive overview of existing methods, evaluate their position within the proposed paradigm, and emphasize the current challenges they face. 
Lastly, we explore potential avenues for future research and identify unresolved issues.

\begin{comment}
The concept of Federated Learning (FL) has garnered increasing attention as a means of conducting collaborative training on decentralized clients while preserving the privacy of individual data.
%
Recent studies have shown that the knowledge of the overall framework is not static, leading to the development of counterpart techniques such as federated increasing learning and federated unlearning. 
%
However, there remains a lack of extensive surveys covering recent advances and thorough analysis of this issue. 
%
In this paper, we present a comprehensive survey on knowledge editing in FL, aiming to summarize the cutting-edge research and broaden the horizons for different domains.
%
Firstly, we propose an integrated paradigm namely Federated Editable Learning (FEL) by rethinking the complete lifecycle of FL.
%
Second, we summarize existing methods, examine their positions in that paradigm, and highlight their current challenges.
%
Finally, we discuss some promising directions and open problems for further research.
\end{comment}
\end{abstract}

\section{Introduction}
\label{Introduction}
%junxiao-2023/1/22
Federated Learning (FL) \cite{mcmahan2017communication} facilitates the collaborative learning of a global model by multiple local clients, while concurrently ensuring secure protection of privacy for each individual client.
It effectively addresses the issue of data silos without completely compromising the privacy of the clients. 
In recent years, FL has garnered significant attention in the academic community and achieved remarkable successes in a variety of industrial applications such as autonomous driving \cite{samarakoon2019distributed}, wearable technology \cite{chen2020fedhealth}, and medical diagnosis \cite{rieke2020future,dayan2021federated}.

In general, the majority of existing FL methods \cite{shoham2019overcoming,wang2021addressing,hong2021federated,yang2021flop} are formulated for static application scenarios, where the data and tasks of the overall FL framework are fixed and known ahead of time. However, in real-world applications, the situation is often dynamic, where local clients receive new task data in an online manner.
To handle this type of situation, researchers are investigating how FL can be adapted to learn continuously over a sequence of tasks.
It has become widely acknowledged that utilizing straightforward stochastic gradient descent (SGD) on FL when learning over a sequence of tasks results in the phenomenon known as ``catastrophic forgetting'', which implies that the model forgets what it had previously learned when acquiring new knowledge \cite{huang2022learn}. 
As a result, a significant proportion of researchers have focused on devising methods namely \emph{federated increasing learning} to augment knowledge while concurrently mitigating the problem of forgetting \cite{dong2022federated}.

On the other hand, forgetting is not always detrimental. Selective amnesia, also referred to as \emph{federated unlearning} \cite{leijie,flquestion,10.1145/3485447.3512222}, which involves the elimination of specific knowledge, can address privacy concerns even create additional ``space'' for acquiring new knowledge. It is possible that in the future, FL will be required to completely remove any indication of having learned a specific data or task.
As we look towards the future, imagine a FL service provider whose system is continuously updated by learning new skills from the data collected from its customers' daily lives. 
Occasionally, the provider may be required to delete previously acquired behaviors and/or knowledge regarding specific tasks or data that have been identified as raising potential fairness \cite{ezzeldin2021fairfed}, privacy \cite{nasr2019comprehensive}, or security concerns \cite{bagdasaryan2020backdoor}.

Nevertheless, there is a scarcity of extensive literature reviews that encompass recent advancements and provide a detailed examination of this subject matter.
As of the time of writing this paper, \emph{federated unlearning} has not yet been well studied in the \emph{federated increasing learning} setting where the underlying data distribution can shift over time.
In this paper, we undertake a comprehensive survey of the field of knowledge editing (augmentation/removal) in FL, with the aim of synthesizing the most recent research advancements and broadening the understanding of its potential applications across various domains.
Overall we make the following contributions:
\begin{itemize}
    \item We introduce an integrated paradigm, referred to as Federated Editable Learning (FEL), by reevaluating the entire lifecycle of FL. 
    \item We present a thorough examination of existing methods, assess their position within the proposed framework, and highlight the current limitations and challenges they encounter.
    \item We investigate the areas for future research and pinpoint unresolved issues.
    Towards efficient lifelong knowledge editing in FL, enabling FL to precisely forget what the user has specified without deteriorating the rest of the acquired knowledge; or, FL to not alter the model's behavior in other contexts when augmenting knowledge.
\end{itemize}

\section{Federated Editable Learning}
\label{Federated Editable Learning}

% Life-cycle of Federated Editable Learning.
In this section, we will introduce the concept of the Federated Learning Lifecycle, including its background, motivation, and definition.
Different from most existing FL applications, we emphasize that a complete FL lifecycle should not only focus on the learning process to obtain a well-trained model, but also empower the reverse unlearning process to ensure user privacy protection.
Therefore, we propose our Federated Editable Learning (FEL) framework to support the sustainable development of a federated learning system.

\begin{figure*}[t]
    \centering
    \includegraphics[width=\linewidth]{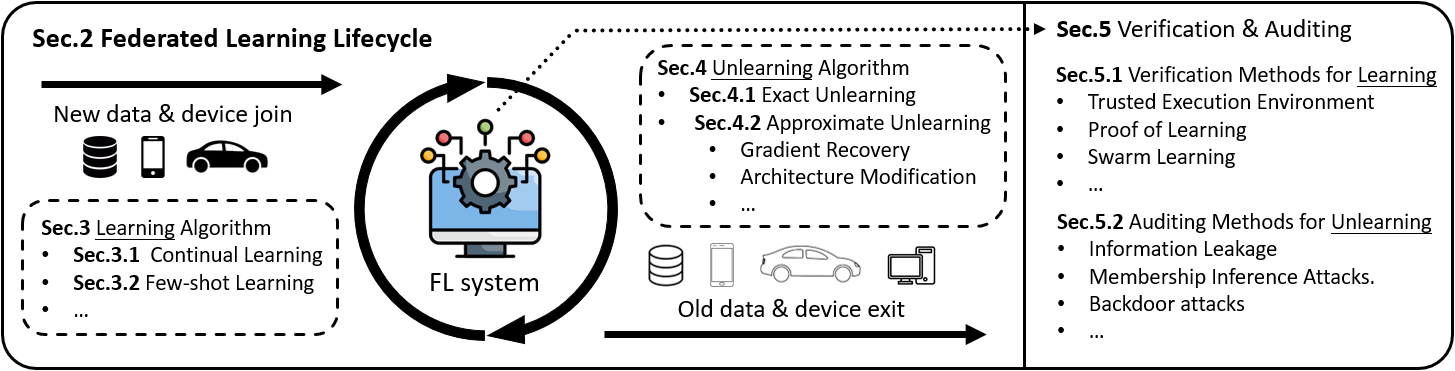}
    \caption{The demonstration of a complete federated learning lifecycle, which defined in Sec.\ref{The Lifecycle of Federated Learning}. The successful operation of FL lifecycle relies on the Federated Editable Learning framework, which consists of two components: the Federated Increasing Learning in Sec.\ref{sec:Federated Increasing Learning} and the Federated Unlearning in Sec.\ref{Federated Unlearning}. Besides, the verification and auditing mechanisms in Sec.\ref{sec:Verification and Auditing} is also important to guarantee the FL lifecycle.}
    \label{fig:FL_lifecycle}
    \vspace{-5pt}
\end{figure*}

\subsection{The Lifecycle of Federated Learning}
\label{The Lifecycle of Federated Learning}
% some background knowledge of FL, such as definitions and workflow (with figures).

Artificial Intelligence (AI) has become an essential component of life today, which achieves significant successes in various domains, such as Computer Vision (CV) \cite{dong2014learning},  Natural Language Processing (NLP), etc \cite{schmidhuber2015deep}.
With the increase in advanced sensing and computing capabilities of ubiquitous mobile devices, AI architectures are gradually shifting from traditional data-centralized cloud server to the distributed edge. 
Besides, considering the importance of user data privacy protection, the federated learning (FL) concept has been proposed.
As an emerging and novel distributed machine learning paradigm, FL adopts collaborative model training on extensive user devices to obtain a model containing globally shared knowledge. Only model parameters are exchanged between the server and user devices, so that the user data never leaves the local side and its privacy protection is guaranteed.
Therefore, involving more user device participation to contribute their data for training is an important and critical principle in the FL scenario.

However, in practical FL applications \cite{li2021survey}, the FL system is always in a dynamic changing process, which can be divided into the following cases:
\begin{itemize}
    \item Data Dynamic: The local data of user devices already involved inside the FL system is constantly updated (generating new data \& deleting obsolete data).
    \item Device Dynamic: The participated user devices of the FL system are also changing (new devices join \& old devices exit). For example, user devices from different time zones have their own available periods.
\end{itemize}
In fact, the essence of both two dynamic cases is all about the data flow in the FL system.
Besides, the current mainstream machine learning models also have their own constraints.
Given a specific model architecture, there is an upper limit on the knowledge amount that the model can contain \cite{roberts2020much}. Generally speaking, a larger model with more parameters can learn more knowledge.

Therefore, for an FL system with given model architecture, the global model has to keep updating its knowledge to adapt the above system dynamic and knowledge constraints.
This model updating involves not only learning new knowledge from new data or devices, but also removing the negative effects of obsolete data or devices from the current model. 
We refer to them as ``Learning process" and ``Unlearning process", respectively.
However, the majority of the existing FL frameworks mainly focus on the learning process, while the unlearning process is neglected \cite{yang2019federated}.
A fixation on only learning new knowledge can lead to the model quickly reaching its knowledge upper limit and thus being unable to further adapt to the system dynamic. 
The ``unlearning process" is also a necessary component of the FL system, which can help remove obsolete knowledge and create space for new knowledge in the future.

Based on the above insights, we are the first to propose the concept of lifecycle for the current FL paradigm, where the demonstration of a complete FL lifecycle is shown in Figure~\ref{fig:FL_lifecycle}.
It incorporates both sides of ``Learning" and ``Unlearning " to achieve the expected model knowledge editing, which enables the sustainable development of the FL system.
% enable the sustainable development of the FL system, and only the organic combination between them can achieve the complete lifecycle of FL systems
In addition, auditing the results of model editing is also a crucial element for the FL lifecycle.
In the learning process, we need to ensure that the user device has performed the corresponding training requirements honestly and credibly.
In the unlearning process, we need to ensure that the knowledge of deleted data is fully removed from the current model, while the knowledge of remaining data is kept unchanged.
%

% However, the majority of the existing FL frameworks mainly focus on the learning process during the FL lifecycle, while the unlearning process is neglected \cite{yang2019federated}.
% %
% A fixation on only learning new knowledge can lead to the model quickly reaching its knowledge upper limit and thus being unable to further adapt to the system dynamic. 
% %
% The ``unlearning process" is also a necessary component for the complete FL lifecycle, which can help remove obsolete knowledge and create space for new knowledge in the future.
% %
% Therefore, there is an urgent demand for the survey of FL lifecycle that needs to be filled.

% one side of the FL lifecycle and ignore the counterpart on the other side.
%

\subsection{What is Federated Editable Learning}
% idea 1: Protect the right of users to \textbf{freely control their data} from two aspects: 1) contribute their data to obtain better model or service or 2) revoke their data to protect privacy.
% category: Federated Increasing Learning and Federated Unlearning
% auditing for 

Carrying the above concept of FL lifecycle, we introduce our framework named Federated Editable Learning (FEL) as the cornerstone for a perfect implementation of the FL lifecycle.
The ultimate objective of FEL is to empower user devices to freely control their own private data in the FL system, while ensuring the FL global model can adaptively adjust its knowledge to handle the system dynamics.
On the one hand, the user devices have the right to decide which part of their data will be contributed to participate in the collaborative training process of FL system, and the knowledge contained in limited data can be absorbed into the FL global model.
On the other hand, the user devices should also have the right to revoke their previously participated data from the FL system, i.e., deleting the historical influences induced by participated data from the current FL global model. 
The model after deletion operation should behave as if these data never participate in FL training, and those relevant obsolete knowledge in the model also need to be removed.
Therefore, to achieve the objectives of FEL in both aspects, we characterize the existing FL-related works into two categories: \textit{Federated Increasing Learning (FIL)} and \textit{Federated Unlearning (FU)}.

\textbf{Federated Increasing Learning (FIL)}: 
The work in this category focuses on how to obtain new knowledge for the global model from the constant data flowing into the FL system, and there are several critical challenges that need to be addressed in FIL.
First, the contributed data of user devices may only occur once in the FL system, the server must leverage the only opportunity to derive knowledge from this single participation.
To address this challenge, we provide a comprehensive survey on \textit{Federated Continual Learning} (FCL), where more details are provided in Sec.\ref{subsec:Federated Continual Learning}.
Second, user devices are constrained by limited resources (e.g., memory space, computation unit, etc), which may result in a very limited amount of data being generated on them. 
As we know, good knowledge representation of AI comes from the big data analysis from massive amounts of data. Thus, how to extract knowledge with generalization from a small amount of specialization data is a serious challenge.
We discover the success of many existing works on \textit{Federated Few-shot Learning} (FFsL) to handle the above challenge, and provide an exhaustive survey to summarize the current research frontier.

\textbf{Federated Unlearning (FU)}: The work in this category focuses on deleting the obsolete data as well as its historical influence in the current model, which not only protects the user data privacy but also creates "space" for new knowledge in the future.
A straightforward way is to retrain a new model from scratch with the remaining data only. 
However, naive retraining demands huge computational resources and time costs, which is completely unacceptable for an FL system. 
% In addition, the system dynamics also make it impossible since some user private data are involved only once in the training history.
%
Therefore, we provide a detailed survey about the existing advanced or optimized retraining-based methods in Sec.\ref{subsec: Exact Federated Unlearning}.
Except for exact retraining, approximate unlearning methods are the mainstream in the current FU field. Its objective is to generate an approximate unlearning model in a fast and computationally efficient manner, whose behavior is almost equivalent to an exact retraining model. A comprehensive survey about approximate FU is provided in Sec.\ref{subsec: Approximate Federated Unlearning}.

\section{Federated Increasing Learning}
\label{sec:Federated Increasing Learning}

In this section, we are going to consider Federated Increasing Learning (FIL) problems, which involve the federated training over time. In the standard FL setting, the objective is to build a joint model using a certain amount of data from a multitude of coordinated devices in a decentralized way. One typical assumption in standard FL is that the whole training dataset is available from the beginning of the training stage. 

However, this assumption rarely holds in real-world FL applications,
% real-world FL applications are often dynamic, 
where local clients often collect new data progressively, during several days, or weeks, depending on the context. Moreover, new clients with unseen new data may participate in the FL training, further aggravating the model and could be unable to converge to a solution. For these reasons, we need to introduce Increasing Learning (IL) \cite{thrun1995lifelong} into FL. FIL research gains a lot of importance since it addresses the difficulties of training a model gradually with data collected over different periods of time, adapting to the new instances and trying to preserve the previous knowledge.

\begin{table*}[t]\normalsize
	\renewcommand\arraystretch{1.2}
	\centering
	\caption{Summary and classification of existing federated increasing learning works.}
        \scalebox{0.93}{
	\begin{tabular}{cc|c|c|c}
        \hline
        \multicolumn{2}{c|}{\textbf{Category}}                                                  & \textbf{Method}  & \textbf{Publication} & \textbf{Key Contribution} \\ \hline
        % \multicolumn{2}{c|}{\textbf{\textit{Exact}}}                                                     &   Cryptography     &  \cite{9514457}           &                  \\ \hline
        \multicolumn{1}{c|}{\multirow{8}{*}{\rotatebox[origin=c]{90}{
        \begin{tabular}[c]{@{}c@{}}\textbf{\textit{FCL}} \end{tabular}
        }}} & \multirow{5}{*}{\begin{tabular}[c]{@{}c@{}} Task-based \\ FCL \end{tabular}} & Ensemble Learning  &     \cite{casado2020federated}           &   Lightweight and real-time framework                \\ \cline{3-5} 
        \multicolumn{1}{l|}{}                             &                            &  Parameter Decomposition &  \cite{yoon2021federated}      &   Selective knowledge aggregation         \\ \cline{3-5}
        \multicolumn{1}{l|}{}                             &              &     Drift Detection      & \cite{casado2022concept}          &   Autonomous user local training strategy          \\ \cline{3-5}
        \multicolumn{1}{l|}{}                             &                            &  Elastic Weight Consolidation      & \cite{shoham2019overcoming}          &  Penalty term to the loss function                \\ \cline{3-5}
        \multicolumn{1}{l|}{}                             &                            &   Elastic Weight Consolidation     & \cite{yao2020continual}          &  Limit updates of important parameter                 \\ 
        % \multicolumn{1}{l|}{}                             &                            &       & 
        % \cite{forget_svgd}          &                  \\ \cline{3-5}
        % \multicolumn{1}{l|}{}                             &                            &        & 
        % \cite{flquestion}          &                  \\ \cline{3-5}
        % \multicolumn{1}{l|}{}                            &   
        %                  &        & \cite{ifu}                 & 
        %               \\ 
        \cline{2-5}
        % \multicolumn{1}{l|}{}                             &                            &        &  \cite{leijie}           &                  \\ \cline{2-5} 
        \multicolumn{1}{l|}{}                             & 
        \multirow{3}{*}{\begin{tabular}[c]{@{}c@{}} Class-based \\ FCL \end{tabular}}               & Regularization Term      & \cite{usmanova2022federated}    & Use distillation loss to transfer knowledge   \\ \cline{3-5}
        % \multicolumn{1}{l|}{}                             &                            &        &   \cite{flquestion}             
        \multicolumn{1}{l|}{}                             &                            &  Prototypical Networks        & \cite{hendryx2021federated}          &   Store exemplars from previous tasks               \\ \cline{3-5}
        \multicolumn{1}{l|}{}                             &                            & Truncated Cross Entropy          & \cite{legate2022reducing}          & Force user learn internal representation                  \\ \cline{2-5}
        \hline
        \multicolumn{1}{c|}{\multirow{5}{*}{\rotatebox[origin=c]{90}{
        \begin{tabular}[c]{@{}c@{}}\textbf{\textit{FFsL}} \end{tabular}
        }}} & \multirow{5}{*}{  \begin{tabular}[c]{@{}c@{}} Generalization  \\ to Unseen Data \end{tabular}    } &  Meta-Learning      & \cite{chen2018federated}           &  First to apply FL on meta-learning                \\ \cline{3-5} 
        \multicolumn{1}{l|}{}                             &                            &  Adversarial Learning      & \cite{fan2021federated}           &  Produce different feature for unseen data                \\ \cline{3-5}
        \multicolumn{1}{l|}{}                             &                            &  Energy-based Weighting        &  \cite{dong2022fewfedweight}          & Updating the weights of pseudo examples                 \\ \cline{3-5}
        \multicolumn{1}{l|}{}                             &                            & Adversarial Learning          & \cite{huang2022few}          &    Latent embedding adaptation              \\ \cline{3-5}
        \multicolumn{1}{l|}{}                             &                            & Client Selection       & \cite{xu2022client}          & Exclude the  malicious user participation                \\ \cline{3-5}
        % \multicolumn{1}{l|}{}                             &                            &        & 
        % \cite{forget_svgd}          &                  \\ \cline{3-5}
        % \multicolumn{1}{l|}{}                             &                            &        & 
        % \cite{flquestion}          &                  \\ \cline{3-5}
        % \multicolumn{1}{l|}{}                             &                            &        & \cite{ifu}          &                  \\ \cline{3-5}
        % \multicolumn{1}{l|}{}                             &                            &       & \cite{leijie}           &                  \\ \cline{2-5} 
        % \multicolumn{1}{l|}{}                             & Architecture               &        & 
        % \cite{10.1145/3485447.3512222}    &                  \\ 
        \hline
        \end{tabular}}
\label{table:fil}
\end{table*}

% some brief introduction to Federated Increasing Learning

\subsection{Challenges in Federated Increasing Learning}

In the FIL setting, each local client collects the training data continuously with its own preference, while new clients with unseen new data could join the FL training at any time. More specifically, the data distributions of the collected classes across the current and newly-added clients are non-independent and identically distributed (non-i.i.d.). FIL requires these local clients to collaboratively train a global model to learn new data continuously, with constraints on privacy preservation and limited memory storage \cite{rebuffi2017icarl,wu2019large}.

Under such circumstances, an ideal framework should recognize new classes and meanwhile maintain discriminability over old classes, which is called Federated Continual Learning (FCL). The main difficulty in FCL is catastrophic forgetting \cite{goodfellow2013empirical}. 
% and concept drift \cite{gama2014survey}. 
Catastrophic forgetting refers to the phenomenon that occurs when optimizing the model with new classes, the formerly acquired knowledge on old classes is quickly forgotten. 
% Concept drift, on the contrary, is a problem that arises when the model is learning a single task, but the statistical properties of the data (what the model is trying to predict) change over time. As a result, the model performance tends to drop dramatically.

Under severe circumstances, only limited novel instances are available to incrementally update the model. Meanwhile, local clients often have very limited storage memory to store few-shot old data. As a result, the task of recognizing few-shot new classes without forgetting old classes is called federated few-shot class-incremental learning. Such lack of data would further exacerbate local forgetting caused by class imbalance at the local clients and global forgetting brought by the non-i.i.d class imbalance across clients.

% limited data and computational resource $\to$ Federated Few-shot Learning
% New data is constantly generated on the user device $\to$ Federated Continual Learning

%-----------------------------------------------------------------
\subsection{Federated Continual Learning}
\label{subsec:Federated Continual Learning}

% In FCL each client has its privately accessible sequence of tasks. Each round client trains its local model on some task from its sequence and then sends its parameters to the server.

Mostly FCL methods address task-continual learning (task-CL) scenario \cite{de2021continual}, where information about the task-ID of examples is known at test time. However, more challenging scenario is class-continual learning (class-CL), where the model has to distinguish among all the classes of all the tasks at test time \cite{masana2020class}. 
In the following part, we will review the literature on task-based FCL and class-based FCL, respectively.

For the task-CL problem in FL, a large number of works focus on the problem of catastrophic forgetting. 
For example, LFedCon2~\cite{casado2020federated} aims to use light, traditional classification models, e.g., a generalized linear model (GLM), a decision tree (DT), to support real-time, continuously and autonomously learning phase in a privacy-preserving and decentralized manner. 
Yoon et al.,~\cite{yoon2021federated} propose FedWeIT, in which each client learns a series of tasks from the private local data stream, meanwhile different clients can also learn from others to enhance their learning performance. Specifically, a learnable mask vector is trained to filter the relevant knowledge from other clients during the aggregation phase.
% based on a decomposition of the model parameters into a dense global parameters and sparse task-adaptive which are shared between all the clients. 
%
% LFedCon2 \cite{casado2020federated} uses traditional classifiers instead of DNN and propose an algorithm dealing with a concept drift based on ensemble retraining.  
%
To solve the concept drift (i.e., the underlying distribution of data can change in unforeseen ways over time), CDA-FedAvg~\cite{casado2022concept} designs a detection mechanism to monitor concept drift, so that each device has enough autonomy to decide when to train and what data to use. By such means, the server will simply orchestrate the process.
% on detecting the distribution shifts in an FCL setting, as ill defined task-IDs presents a fresh set of challenges if they are not well specified at training time. 
FedCurv \cite{shoham2019overcoming} and FedCL \cite{yao2020continual} adopt EWC \cite{kirkpatrick2017overcoming}, which aims to improve the generalization ability of the federated models.

For class-CL scenarios in FL, FCL methods can be divided into the following three types \cite{masana2020class}: 
\begin{itemize}
    \item [1)] \textbf{Regularization-based approaches},  which compute the importance of weights for previous tasks and penalize the model for changing them (i.e., FLwF~\cite{usmanova2022federated} use distillation loss to transfer the knowledge from the server and decrease the forgetting of previously learned tasks); 
    \item [2)] \textbf{Exemplar-based approaches},  which store exemplars from previous tasks, i.e., Hendryx et al.,~\cite{hendryx2021federated} use federated prototypical networks to efficiently learn new classes in sequence;
    \item[3)] \textbf{Bias-correction approaches},  which deal explicitly with bias towards recently-learned tasks (Legate et al.,~\cite{legate2022reducing} adopt Truncated Cross Entropy (TCE) to force each client to learn by adapting the model’s internal representation of the classes present in its training data). 
\end{itemize}

%-----------------------------------------------------------------
\subsection{Federated Few-shot Learning}
% Training FSL models on distributed devices is still an under-explored open problem. Few-Shot Learning aims to extract the inductive bias from base classes and generalize it to unseen classes. The first work of this topic was from Chen et al.,~\cite{chen2018federated} who explored federated meta-learning by applying FedAvg on meta-learning approaches such as MAML~\cite{finn2017model} in a straightforward way.
With the goal of extracting the inductive bias from base classes and generalizing it to unseen classes, few-shot learning has been widely explored in recent years. However, training FSL models on distributed devices is still an open problem. 
% The existing approaches for FFsL can be grouped into the following categories: data-based, model-based, and algorithm-based methods.
%
%
% \subsubsection{Data-based FFsL}
% The first work of this topic was from Chen et al.,~\cite{chen2018federated} who explored federated meta-learning by applying FedAvg on meta-learning approaches such as MAML~\cite{finn2017model} in a straightforward way. 
%
% The core issue of FFsL is the unreliability of the empirical risk minimizer, due to limited training samples~\cite{wang2020generalizing}. 
% The existing approaches for FFsL can be grouped into the following categories: data-based, model-based, and algorithm-based.
%
% Therefore, some researchers propose data augmentation-based FFsL methods to reduce the uncertainty of the empirical risk minimizer by enriching supervised information, e.g., Aug-FedPrompt~\cite{cai2022aug} defines a data generator that carefully annotates abundant unlabeled data for data augmentation; 
%
%
The first work of this topic was from Chen et al.,~\cite{chen2018federated} who explored federated meta-learning by applying FedAvg on meta-learning approaches such as MAML~\cite{finn2017model} in a straightforward way. 
Another line of work focuses on data augmentation to alleviate data scarcity, i.e., FewFedWeight~\cite{dong2022fewfedweight} proposes an energy-based weighting algorithm for updating the weights of pseudo examples generated by the global model and a dynamic aggregation method based on the performance of client models.
% ; Fed-ZDA~\cite{hao2021towards} generates synthetic data of the desired classes by finding the data that results in similar statistics as those stored in the batch normalization (BN) layers of the pretrained model.
%
% \subsubsection{Model-based FFsL}
% Model-based FFsL methods constrain the hypothesis space via prior knowledge to reduce overfitting and make the empirical risk minimizer more reliable. To obtain an efficient few-shot model, adversarial learning, 
%
Then, FedFSL~\cite{fan2021federated} formulates the training in an adversarial fashion and optimizes the client models to produce a discriminative feature space that can better represent unseen data samples, while FedAffect~\cite{huang2022few} considers a more challenge scenario: local participants design their models independently.
%
% FedProto~\cite{tan2022fedproto} proposes a novel federated prototype learning (FedProto) framework in which the clients and server communicate the abstract class prototypes instead of the gradients. 
%
% FEWFEDWEIGHT 
% Dong et al.,~\cite{dong2022fewfedweight} proposes
% an energy-based weighting algorithm for updating the weights of pseudo examples generated by the global model and a dynamic aggregation method based on the performance of client models.
%
% Huang et al.,~\cite{huang2022few} investigate a novel and challenging problem (FSFL): model-agnostic federated learning with a small number of samples,
% of a small sample of model-independent Federated Learning, 
% known as FSFL(little-shot Model Agnostic Federated Learning), 
% in which local participants design their independent models from a limited number of private data set. Given the scarcity of private data, the authors leverage a rich set of publicly available data to bridge the gap between local private participants.
% To address the domain gap between public and private domain, they 
% Besides, an adversarial learning scheme is proposed to discriminate the gap between public and private domains. 
% with an adversarial learning scheme
% propose a novel framework based on latent embedding adaptation under a public-private communication manner.
%
% Shome et al., ~\cite{shome2021fedaffect,sun2022lightweight}introduce Relation Network~\cite{sung2018learning} in FL for industrial scenarios and facial expression recognition.
%
% \subsubsection{Algorithm-based FFsL}
Besides that, 
% With prior knowledge, algorithm-based FFsL methods seek optimal global model by directly learning an optimizer to output search steps or by providing a good aggregation algorithm. 
%
%
%
CSFedL \cite{xu2022client} proposes an adaptive client selection strategy to mitigate the impact caused by malicious participation, to obtain a more effective few-shot model.

% Few-Shot Learning aims to extract the inductive bias from base classes and generalize it to unseen classes. 
% Current FedFSL methods can also be roughly divided into two groups: optimization-based methods and metric-based methods.

\section{Federated Unlearning}
\label{Federated Unlearning}

After the FL training of a model is completed, clients may require the FL server to remove parts of data contribution from the global model to protect the user’s privacy and avoid legal risks.
The scenario is called \emph{federated unlearning}.
The server should transform the model into an updated one that operates as if those deleted data never participated in FL training. 
% The mathematical formulation of the federated unlearning problem is given as

% \begin{definition}[Federated Unlearning]
%     We define the algorithm $\mathcal{L}$ to be a learning process that maps ... We define an unlearning process $\mathcal{L}^-$ ...
% \end{definition}

In the section, we first discuss several major challenges in federated unlearning and then summarize the emerging federated unlearning works from perspectives of \emph{exact federated unlearning} and \emph{approximate federated unlearning}.

\begin{table*}[t]\normalsize
	\renewcommand\arraystretch{1.23}
	\centering
	\caption{Summary and classification of existing federated unlearning works.}
        \scalebox{0.95}{
	\begin{tabular}{cc|c|c|c|c}
        \hline
        \multicolumn{2}{c|}{\textbf{Category}}                                                  & \textbf{Method} & \textbf{Scenario} & \textbf{Publication} & \textbf{Unlearning Requests} \\ \hline
        \multicolumn{2}{c|}{\multirow{3}{*}{\textbf{\textit{Exact}}}}                                                     &   Ensemble  & DNN  &  \cite{ijcai2022p556}           &  Client, Class, Sample   \\ \cline{3-6}
                      &                            &    Ensemble  &  DNN  &  \cite{yu2022legonet}           &      Client, Class, Sample            \\ \cline{3-6}
                      &                            &   Cryptography  & Random Forest  &  \cite{9796721}           &         Client         \\ \hline
        \multicolumn{1}{c|}{\multirow{12}{*}{\rotatebox[origin=c]{90}{\textbf{\textit{Approximate}}}}} & \multirow{4}{*}{ \begin{tabular}[c]{@{}c@{}} Gradient \\ Recovery \end{tabular} } &   History Retaining  &  DNN  &  \cite{9521274}           &       Client           \\ \cline{3-6} 
        \multicolumn{1}{l|}{}                             &                            & History Retaining  &  DNN   &  \cite{9796721}           &      Client            \\ \cline{3-6}
        \multicolumn{1}{l|}{}                             &                            &  History Retaining  &  Poisoning Recovery  &   \cite{fedrecover}          &    Client              \\ \cline{3-6}
        \multicolumn{1}{l|}{}                             &                             &  History Retaining  &   Recommender System  &   \cite{recommendation}          &      Client            \\ \cline{2-6}
        \multicolumn{1}{l|}{}                             &   \multirow{4}{*}{\begin{tabular}[c]{@{}c@{}} Parameter \\ Updating \end{tabular}}    &    Knowledge Distillation    &  DNN  &\cite{kdunlearning}          &    Client              \\ \cline{3-6}
        \multicolumn{1}{l|}{}                             &                            &  Gradient Descent      & Bayesian Model  &\cite{forget_svgd}          &        Client          \\ \cline{3-6}
        \multicolumn{1}{l|}{}                             &                            &  Gradient Descent      &  DNN  &\cite{flquestion}          &      Client            \\ \cline{3-6}
        \multicolumn{1}{l|}{}                             &                            &  Gradient Descent    &  DNN &\cite{leijie}           &   Client, Class, Sample               \\ \cline{2-6} 
        \multicolumn{1}{l|}{}                             & \multirow{2}{*}{\begin{tabular}[c]{@{}c@{}} Architecture \\ Modification \end{tabular}}    &  Channel Pruning      &  DNN  &\cite{10.1145/3485447.3512222}    &          Class        \\ \cline{3-6}
        \multicolumn{1}{l|}{}      &    &  Output Filtering  &  DNN  & \cite{baumhauer2022machine}    &     Class             \\ \cline{2-6}
        \multicolumn{1}{l|}{}      &  \multirow{2}{*}{\begin{tabular}[c]{@{}c@{}} Noise \\ Perturbation \end{tabular}}   &  Differential Privacy   & DNN &   \cite{gupta2021adaptive}  &      Client, Class, Sample          \\   \cline{3-6}
        \multicolumn{1}{l|}{}      &   & Randomized Perturbation   & DNN  & \cite{ifu}    &     Client             \\ \hline
        \end{tabular}}
\label{table:contract}
\end{table*}

\subsection{Challenges in Federated Unlearning}

Compared with traditional machine learning, the characteristics of FL bring three major challenges to the unlearning technique as follows.

\paragraph{1) Iterative Learning:} At the beginning of each round in FL, the model of each client is the aggregation result for all clients in the previous round. 
Such an intertwining of client training results in each round leads to the fundamental challenge of federated unlearning.

\paragraph{2) Information Isolation:} Privacy protection, one of the major advantages of FL, prevent FL servers from accessing the client data. 
In other words, every client maintains its data samples and trains the model locally. 

\paragraph{3) Stochastic Training:} The process of FL training is non-deterministic. 
For each round, the FL server randomly selects the clients for global aggregation while each client randomly selects and orders batches of data for local training.

%-----------------------------------------------------------------
\subsection{Exact Federated Unlearning}
\label{subsec: Exact Federated Unlearning}

A naive way to make the best FL model that provably forgets the target data is to retrain a new model based on the remaining data from scratch. 
However, it is prohibitively expensive for an FL server to fully retrain a model in terms of computation and time overhead.
Some works are designed to achieve the unlearning in such a way that the produced models are effectively the same as the ones obtained with retraining but at a cheaper computing cost.
We call these works exact federated unlearning, which are summarized as follows.
% Moreover, the FL clients do not have the incentive to participate in the retraining.

Some ensemble learning-based works are designed for machine unlearning originally, however, their idea can be applied in federated unlearning.
% when relaxing the assumption of privacy protection in FL.
For example, 
% \cite{9519428} ...
\cite{ijcai2022p556} propose an efficient exact unlearning framework.
It divides the dataset into several isolated sub-datasets, each corresponding to a sub-model, accelerating the retraining process and ensuring the retrained model's accuracy.
\cite{yu2022legonet} present a novel neural network named LegoNet, composed of a fixed encoder (i.e., the backbone for representation learning) and multiple isolated adapters to be retrained for unlearning.
The adapters occupy few parameters of LegoNet; thus, the re-trained parameters during unlearning can be significantly reduced.
% TODO: how they can be applied in FL ...

Moreover, without compromising the privacy of clients in FL, \cite{9514457} develop a cryptography-based approach for federated unlearning.
It presents a revocable federated learning framework for random forest (RF) called RevFRF by designing a customised homomorphic encryption-based protocol. 
RevFRF guarantees two levels of unlearning: 1) the remaining participants cannot utilize the data of an honest and leaving participant in the trained model; 
2) a dishonest participant cannot get back to utilize the data of the remaining participants memorized by  the trained model.

%-----------------------------------------------------------------
\subsection{Approximate Federated Unlearning}
\label{subsec: Approximate Federated Unlearning}

Although the existing works for exact federated unlearning alleviate the expense of retraining to some extent, their cost is still unacceptable in most FL applications.
Thus, recent works achieve higher efficiency of federated unlearning by relaxing the effectiveness and certifiability requirements for the new model after unlearning, which is called approximate federated learning.
% The precise theoretical definition of approximate federated learning is:
% \begin{definition}[\textit{$\epsilon$-Approximate Unlearning}]
% \label{definition_1}
%     Assume we have an original dataset $\mathcal{D}$, a learning algorithm $\mathcal{A}$, a deleted sub-dataset $\mathcal{D}_f$, and an unlearning algorithm $\mathcal{U}$. 
%     Given $\epsilon > 0$, an approximate federated unlearning algorithm $\mathcal{U}$ is said to achieve $\epsilon$-bounded data deletion for $\mathcal{D}_f$ on a well-trained model $\omega_{\mathcal{D}}$ if 
%     \begin{equation}
%         e^{-\epsilon} \leq \frac{Pr(\mathcal{U} \left( \omega_{\mathcal{D}}, \mathcal{D}_f, \mathcal{D} \right))}{Pr(\mathcal{A}(\omega, \mathcal{D}\backslash \mathcal{D}_f))} \leq e^{\epsilon},
%     \end{equation}
%     where $Pr(\cdot)$ denotes the model distribution of all possible models generated from the specific algorithm, and $\mathcal{A}(\omega, \mathcal{D}\backslash \mathcal{D}_f)$ denotes to retrain a new model from the initial model $\omega$ on the remaining dataset $\mathcal{D}\backslash \mathcal{D}_f$.
% \end{definition}
% The existing work has attempted to achieve the above theoretical objectives from different perspectives as follows.

\subsubsection{Gradient Recovery}
\label{sec:federated_unlearning_retrain}

To overcome the high resource cost caused by the model retraining of the exact federated unlearning described in Sec.\ref{subsec: Exact Federated Unlearning}, the gradient recovery-based approach reconstructs the unlearned model based on the historical parameter updates of clients that have been retained at the FL server during the training process.

\cite{9521274} propose the first federated unlearning approach named FedEraser, reconstructing a new model based on the historical parameter updates of clients stored in the FL server.
To speed up the retraining while maintaining the model performance, FedEraser has a calibration method for the stored historical updates.
\cite{9796721} propose an efficient retraining algorithm based on the diagonal empirical Fisher Information Matrix (FIM) for FL, by observing the first-order Taylor expansion of the loss function during the unlearning process. 
Moreover, to reduce approximation errors in retraining, the proposed algorithm has an adaptive momentum technique.

% Besides the previous works for federated unlearning..., there are some works ... as follows.
\cite{fedrecover} propose an FL model recovery method to recover a model from poisoning attacks using historical information rather than training from scratch.
For each recovery, the server can estimate the model update of a client in each round based on its stored historical information during the past training process.
\cite{recommendation} propose a federated recommendation unlearning method tailed for FL-based recommendation systems (FedRecs). 
The main idea is to revise historical updates and leverage the revised updates to speed up the reconstruction of a FedRec. 

\subsubsection{Parameter Updating}

The above works for federated unlearning via gradient recovery require the FL server to store historical updates, which burdens the server. 
Therefore, another group of federated unlearning is to scrub the trained FL model of information to be forgotten, which we summarize as follows.

\cite{kdunlearning} propose a federated unlearning method to eliminate a client’s contribution by subtracting the accumulated historical updates from the model and leveraging the knowledge distillation method to restore the model’s performance without using any data from the clients.
\cite{forget_svgd} develop a Bayesian federated unlearning method called Forget-Stein Variational Gradient Descent (Forget-SVGD) based on SVGD, a particle-based approximate Bayesian inference approach via gradient-based deterministic updates.
\cite{flquestion} allow a client to perform the unlearning by training a model to maximize the empirical loss via a Projected Gradient Descent algorithm.

The previous works mainly focus on client-level federated unlearning (i.e., removing the data of a specific client from the model).
To solve the limitation, \cite{leijie} propose a general framework covering client-level, class-level, and sample-level federated unlearning. 
The framework comprises a reverse stochastic gradient ascent (SGA) algorithm with elastic weight consolidation (EWC) to achieve fine-grained elimination of training data at different levels. 

\subsubsection{Architecture Modification}

Some approaches implement federated unlearning by modifying the model architecture.
For example, \cite{10.1145/3485447.3512222} propose a channel pruning-based method to remove information about particular classes in an FL model. 
Its main idea is to quantify the class information learned by each channel without globally accessing the data, and then forget special classes by pruning the channels with the most class  discrimination.
\cite{baumhauer2022machine} propose an output filtering technique to remove particular classes in logit-based classification models by applying linear transformation to the output logits, but do not modify the weights in the models.

\subsubsection{Noise Perturbation}

% ...
\cite{gupta2021adaptive} focuses on randomly perturbing the trained model to unlearn specific data samples, which is motivated by the idea of differential privacy.
\cite{ifu} propose a new federated unlearning scheme named \emph{informed federated unlearning} that unlearns a client's contribution with quantifiable unlearning guarantees. 
Unlearning guarantees are provided by introducing a randomized mechanism to perturb an intermediate model selected from the training process with client-specific noise.

\section{Verification \& Auditing}
\label{sec:Verification and Auditing}

The previous literature review summarizes the state-of-the-art approaches for knowledge editing in the FL scenario, which together serve as a support to realize the FL system lifecycle.
In addition, to guarantee our objectives are fully achieved according to specified requirements during the FL lifecycle, the verification for the learning process and auditing for the unlearning process are also critical system components, where a comprehensive survey on them is provided in this section.

%-----------------------------------------------------------------

\subsection{Verification Methods for Learning Process}
Although the increasing learning process can enable the Fl global model always to acquire new knowledge from the data to adapt the system dynamics, the expected knowledge can only be obtained by correct training according to the specified requirements. 
Thus, the learning process of user devices must be verifiable to ensure knowledge correctness.
We summarize several tools able to verify the FL process as follows.

\subsubsection{Trusted Execution Environment}

As a secure environment maintained by each CPU, Trusted Execution Environment (TEE) is a hardware technology that outsources code execution on a protected memory region named \emph{enclave} in any untrusted devices and enables the verification of the execution results. 
Some existing TEE-based FL frameworks depend on the TEE deployment on FL servers and user devices~\cite{9155414}. 
Similarly, a verifiable FIL framework can be realized by outsourcing the increasing learning process to the user devices' TEE.

\subsubsection{Proof of Learning}

The concept of Proof of Learning (PoL) proposed by \cite{9519402} enables a verifier (e.g., an FL server) to assess the integrity of training computations for untrusted workers (e.g., user devices). 
Its main idea is to verify if a sequence of intermediate states (i.e., checkpoints of intermediate weights) came from training and are not random (or worse, forged by a malicious party). 
To prove it, the workers need to provide a sequence of batch indices for the same intermediate model updates.
Although the PoL is a general approach, it may leak the privacy of user devices in FL, which remains to be solved.

\subsubsection{Swarm Learning}

For auditing and accountability in FIL, it is necessary to record the increasing learning process in a public, transparent, and tamper-proof ledger.
By combining blockchain technology with FL for a new learning scheme called \emph{swarm learning}~\cite{warnat2021swarm}, the learning process can be fully recorded by such a ledger in a distributed manner despite the existence of malicious user devices.

%-----------------------------------------------------------------
\subsection{Auditing Methods for Unlearning Process}
It's easy to understand that the auditing mechanism is unnecessary in the category of ``exact" FU, such as retraining since the revoked data is never involved in the new retraining model.
However, for the another mainstream of ``approximate" FU category, the auditing mechanism is critical and necessary, which can validate the effectiveness of these methods, i.e., How large is the difference between the approximate model and the exact retraining model?

\textbf{Membership inference attack}: Given a data sample and the black-box access of the trained model, the goal of this kind of attack is to detect whether the data sample is inside the training dataset of this model \cite{shokri2017membership}.
More specifically, we use adversarial machine learning to obtain an inference model, which can recognize the differences in the target model's prediction results on the inputs that inside the training dataset versus outside the training dataset \cite{nasr2019comprehensive,hu2021source}.
Membership inference attack is very effective on detecting data leakage, which can reflect whether the approximate unlearning still contains the information of deleted data or not.

\textbf{Information Leakage}: Many machine learning models will inherently leak some private information during their training process \cite{pustozerova2020information}, such as the intermediate gradients . 
Many existing works have shown that the raw training data can be recovered with the gradients of each update step, where the gradients are accessible for both user devices and server in the FL scenario. 
This kind of information leakage is utilized and called gradient inversion attacks \cite{zhang2022survey}.
Therefore, we can compare the model difference before and after unlearning to infer whether the information of deleted data will still be leaked.

\textbf{Backdoor attacks}: The techniques of backdoor attack are proposed to inject backdoors into the training data samples to poison the model. The derived model will make accurate predictions on clean data samples, but trigger the backdoor to make the wrong predictions on contaminated data samples.
The backdoor attack technique can be utilized to validate the effectiveness of approximate FU.
More specifically, the user devices can contaminate part of their own data samples during the FL training process \cite{sun2019can,bagdasaryan2020backdoor,wang2020attack}. 
If the contaminated data samples are successfully deleted, the unlearning model will predict them into their correct class. Otherwise, the unlearning model will trigger the backdoor to assign them to the wrong class.

% \cite{verifi} grant the leaving participant the right to verify (RTV) based on a unified verification mechanism including two key verification steps, i.e., marking and checking.
% The main idea is that the marking step selects and tags training examples as markers and then the checking step verifies the degree of unlearning based on different verification metrics defined. 
% Benefiting from RTV, the leaving participant can actively verify the unlearning effect in the next few communication rounds.

%-----------------------------------------------------------------
\section{Conclusion \& Future Vision}

In this paper, we introduce a novel concept of the FL lifecycle, which integrates both federated increasing learning (FIL) and federated unlearning (FU) to achieve knowledge editing for the FL system.
As far as we know, it is the first time providing a comprehensive survey on FL system knowledge editing, including concepts, perspectives, challenges, and future vision.
We summarize the state-of-the-art approaches for knowledge learning \& unlearning in FL system and organize a clear taxonomy of them for handling different challenges.
Moreover, we reclassify the representative verification and auditing mechanisms, which ensure that the knowledge editing process follows the specified requirements while the results are consistent with the expectation.
Although each of the current knowledge editing techniques achieves similar purposes, they are still independent of each other because of the different methods used. 
% There is an urgent need for a self-contained and unified knowledge editing framework that can flexibly achieve both learning and unlearning requirements. 
Therefore, we discuss some promising directions within the future vision.

\textbf{Flexible \& Unified Knowledge Editing}:
So far, knowledge editing needs to find respective solutions for various challenges, which makes their application scenarios very limited. 
There is an urgent need for a self-contained and unified knowledge editing framework that can flexibly achieve both learning and unlearning requirements. 
For example, gradient descent is applied for the learning process, while gradient ascent may be utilized for the unlearning process. 
A unified framework allows for more freedom of knowledge editing within the system and a large number of subsequent derivative efforts based on the same technical kernel can be integrated into the framework as components.

\textbf{Knowledge Disentanglement \& Reassembly}:
Another promising direction is knowledge architecture advances. If the knowledge architecture inside the model can achieve free assembly like building blocks, the whole knowledge editing process will become extremely easy.
A few existing works have designed multi-branch models for knowledge disentanglement in FL systems, which enables the user devices to independently extract different knowledge representations. 
For example, the whole model knowledge of each user can be disentangled into global shared knowledge and local personalized knowledge in \cite{luo2022disentangled}.
Therefore, the learning and unlearning processes just need to simply add or remove the corresponding branches.

\textbf{Trustworthy FL Community}: 
With the explosive growth of user devices, user heterogeneity, and their varying relationships, it's difficult for traditional FL with an authoritative server to manage the whole community. The ultimate future form of FL is an autonomous and trustworthy community with massive participants, where blockchain-based techniques are the foundation to support this future vision.
Each user's activity details in the FL community are uploaded to their respective blocks for maintenance, including the learning and unlearning records, the data usage (only the data index, not the local raw data itself), the resource allocation, and so on.
Any participants in the community can initiate the verification and auditing for others.

\bibliographystyle{named}
\bibliography{ijcai22}

\begin{thebibliography}{}

\bibitem[\protect\citeauthoryear{Bagdasaryan and et
  al}{2020}]{bagdasaryan2020backdoor}
Eugene Bagdasaryan and et~al.
\newblock How to backdoor federated learning.
\newblock In {\em ICAIS}, pages 2938--2948. PMLR, 2020.

\bibitem[\protect\citeauthoryear{Baumhauer and {et
  al}}{2022}]{baumhauer2022machine}
Thomas Baumhauer and {et al}.
\newblock Machine unlearning: Linear filtration for logit-based classifiers.
\newblock {\em Machine Learning}, 111(9):3203--3226, 2022.

\bibitem[\protect\citeauthoryear{Cao and et al}{2023}]{fedrecover}
X.~Cao and et~al.
\newblock Fedrecover: Recovering from poisoning attacks in federated learning
  using historical information.
\newblock In {\em IEEE Symposium on Security and Privacy (SP)}, pages 326--343,
  2023.

\bibitem[\protect\citeauthoryear{Casado and et al}{2020}]{casado2020federated}
Fernando~E Casado and et~al.
\newblock Federated and continual learning for classification tasks in a
  society of devices.
\newblock {\em arXiv:2006.07129}, 2020.

\bibitem[\protect\citeauthoryear{Casado and et al}{2022}]{casado2022concept}
Fernando~E Casado and et~al.
\newblock Concept drift detection and adaptation for federated and continual
  learning.
\newblock {\em Multimedia Tools and Applications}, 81(3):3397--3419, 2022.

\bibitem[\protect\citeauthoryear{Chen and et al}{2018}]{chen2018federated}
Fei Chen and et~al.
\newblock Federated meta-learning with fast convergence and efficient
  communication.
\newblock {\em arXiv preprint arXiv:1802.07876}, 2018.

\bibitem[\protect\citeauthoryear{Chen and et al}{2020}]{chen2020fedhealth}
Yiqiang Chen and et~al.
\newblock Fedhealth: A federated transfer learning framework for wearable
  healthcare.
\newblock {\em IEEE Intelligent Systems}, 35(4):83--93, 2020.

\bibitem[\protect\citeauthoryear{Dayan and et al}{2021}]{dayan2021federated}
Ittai Dayan and et~al.
\newblock Federated learning for predicting clinical outcomes in patients with
  covid-19.
\newblock {\em Nature medicine}, 27(10):1735--1743, 2021.

\bibitem[\protect\citeauthoryear{De~Lange and et al}{2021}]{de2021continual}
Matthias De~Lange and et~al.
\newblock A continual learning survey: Defying forgetting in classification
  tasks.
\newblock {\em IEEE TPAMI}, 44(7):3366--3385, 2021.

\bibitem[\protect\citeauthoryear{Dong and et al}{2014}]{dong2014learning}
Chao Dong and et~al.
\newblock Learning a deep convolutional network for image super-resolution.
\newblock In {\em ECCV}, pages 184--199. Springer, 2014.

\bibitem[\protect\citeauthoryear{Dong and et al}{2022}]{dong2022fewfedweight}
Weilong Dong and et~al.
\newblock Fewfedweight: Few-shot federated learning framework across multiple
  nlp tasks.
\newblock {\em arXiv:2212.08354}, 2022.

\bibitem[\protect\citeauthoryear{Dong \bgroup \em et al.\egroup
  }{2022}]{dong2022federated}
Jiahua Dong, Lixu Wang, Zhen Fang, Gan Sun, Shichao Xu, Xiao Wang, and Qi~Zhu.
\newblock Federated class-incremental learning.
\newblock In {\em CVPR}, pages 10164--10173, 2022.

\bibitem[\protect\citeauthoryear{Ezzeldin and et
  al}{2021}]{ezzeldin2021fairfed}
Yahya~H Ezzeldin and et~al.
\newblock Fairfed: Enabling group fairness in federated learning.
\newblock {\em arXiv preprint arXiv:2110.00857}, 2021.

\bibitem[\protect\citeauthoryear{Fan and et al}{2021}]{fan2021federated}
Chenyou Fan and et~al.
\newblock Federated few-shot learning with adversarial learning.
\newblock In {\em WiOpt}, pages 1--8. IEEE, 2021.

\bibitem[\protect\citeauthoryear{Finn and et al}{2017}]{finn2017model}
Chelsea Finn and et~al.
\newblock Model-agnostic meta-learning for fast adaptation of deep networks.
\newblock In {\em ICML}, pages 1126--1135. PMLR, 2017.

\bibitem[\protect\citeauthoryear{Fraboni and et al}{2022}]{ifu}
Yann Fraboni and et~al.
\newblock Sequential informed federated unlearning: Efficient and provable
  client unlearning in federated optimization, 2022.

\bibitem[\protect\citeauthoryear{Gong and et al}{2022}]{forget_svgd}
Jinu Gong and et~al.
\newblock Forget-svgd: Particle-based bayesian federated unlearning.
\newblock In {\em IEEE DSLW}, pages 1--6, 2022.

\bibitem[\protect\citeauthoryear{Goodfellow and et
  al}{2013}]{goodfellow2013empirical}
Ian~J Goodfellow and et~al.
\newblock An empirical investigation of catastrophic forgetting in
  gradient-based neural networks.
\newblock {\em arXiv:1312.6211}, 2013.

\bibitem[\protect\citeauthoryear{Gupta \bgroup \em et al.\egroup
  }{2021}]{gupta2021adaptive}
Varun Gupta, Christopher Jung, Seth Neel, Aaron Roth, Saeed~Sharifi Malvajerdi,
  and Christopher Waites.
\newblock Adaptive machine unlearning.
\newblock In A.~Beygelzimer, Y.~Dauphin, P.~Liang, and J.~Wortman Vaughan,
  editors, {\em Advances in Neural Information Processing Systems}, 2021.

\bibitem[\protect\citeauthoryear{Halimi and et al}{2022}]{flquestion}
Anisa Halimi and et~al.
\newblock Federated unlearning: How to efficiently erase a client in fl?
\newblock In {\em ICML}, 2022.

\bibitem[\protect\citeauthoryear{Hendryx and et
  al}{2021}]{hendryx2021federated}
Sean~M Hendryx and et~al.
\newblock Federated reconnaissance: Efficient, distributed, class-incremental
  learning.
\newblock {\em arXiv:2109.00150}, 2021.

\bibitem[\protect\citeauthoryear{Hong and et al}{2021}]{hong2021federated}
Junyuan Hong and et~al.
\newblock Federated adversarial debiasing for fair and transferable
  representations.
\newblock In {\em ACM SIGKDD}, pages 617--627, 2021.

\bibitem[\protect\citeauthoryear{Hu and et al}{2021}]{hu2021source}
Hongsheng Hu and et~al.
\newblock Source inference attacks in federated learning.
\newblock In {\em 2021 IEEE International Conference on Data Mining (ICDM)},
  pages 1102--1107. IEEE, 2021.

\bibitem[\protect\citeauthoryear{Huang and et al}{2022a}]{huang2022few}
Wenke Huang and et~al.
\newblock Few-shot model agnostic federated learning.
\newblock In {\em ACM MM}, pages 7309--7316, 2022.

\bibitem[\protect\citeauthoryear{Huang and et al}{2022b}]{huang2022learn}
Wenke Huang and et~al.
\newblock Learn from others and be yourself in heterogeneous federated
  learning.
\newblock In {\em CVPR}, pages 10143--10153, 2022.

\bibitem[\protect\citeauthoryear{Jia and et al}{2021}]{9519402}
Hengrui Jia and et~al.
\newblock Proof-of-learning: Definitions and practice.
\newblock In {\em IEEE Symposium on Security and Privacy (SP)}, pages
  1039--1056, 2021.

\bibitem[\protect\citeauthoryear{Kirkpatrick and et
  al}{2017}]{kirkpatrick2017overcoming}
James Kirkpatrick and et~al.
\newblock Overcoming catastrophic forgetting in neural networks.
\newblock {\em Proceedings of the national academy of sciences},
  114(13):3521--3526, 2017.

\bibitem[\protect\citeauthoryear{Legate and et al}{2022}]{legate2022reducing}
Gwen Legate and et~al.
\newblock Reducing forgetting in federated learning with truncated
  cross-entropy.
\newblock In {\em NeurIPS 2022 Workshop}, 2022.

\bibitem[\protect\citeauthoryear{Li and et al}{}]{li2021survey}
Qinbin Li and et~al.
\newblock A survey on federated learning systems: vision, hype and reality for
  data privacy and protection.
\newblock {\em IEEE TKDE}.

\bibitem[\protect\citeauthoryear{Liu and et al}{2021}]{9521274}
Gaoyang Liu and et~al.
\newblock Federaser: Enabling efficient client-level data removal from
  federated learning models.
\newblock In {\em IEEE/ACM IWQOS}, pages 1--10, 2021.

\bibitem[\protect\citeauthoryear{Liu and et al}{2022a}]{9514457}
Yang Liu and et~al.
\newblock Revfrf: Enabling cross-domain random forest training with revocable
  federated learning.
\newblock {\em IEEE TDSC}, 19(6):3671--3685, 2022.

\bibitem[\protect\citeauthoryear{Liu and et al}{2022b}]{9796721}
Yi~Liu and et~al.
\newblock The right to be forgotten in federated learning: An efficient
  realization with rapid retraining.
\newblock In {\em IEEE INFOCOM}, pages 1749--1758, 2022.

\bibitem[\protect\citeauthoryear{Luo and et al}{2022}]{luo2022disentangled}
Zhengquan Luo and et~al.
\newblock Disentangled federated learning for tackling attributes skew via
  invariant aggregation and diversity transferring.
\newblock {\em arXiv preprint arXiv:2206.06818}, 2022.

\bibitem[\protect\citeauthoryear{Masana and et al}{2020}]{masana2020class}
Marc Masana and et~al.
\newblock Class-incremental learning: survey and performance evaluation on
  image classification.
\newblock {\em arXiv:2010.15277}, 2020.

\bibitem[\protect\citeauthoryear{McMahan and et
  al}{2017}]{mcmahan2017communication}
Brendan McMahan and et~al.
\newblock Communication-efficient learning of deep networks from decentralized
  data.
\newblock In {\em Artificial intelligence and statistics}, pages 1273--1282.
  PMLR, 2017.

\bibitem[\protect\citeauthoryear{Nasr and et al}{2019}]{nasr2019comprehensive}
Milad Nasr and et~al.
\newblock Comprehensive privacy analysis of deep learning: Passive and active
  white-box inference attacks against centralized and federated learning.
\newblock In {\em IEEE symposium on security and privacy (SP)}, pages 739--753,
  2019.

\bibitem[\protect\citeauthoryear{Pustozerova and et
  al}{2020}]{pustozerova2020information}
Anastasia Pustozerova and et~al.
\newblock Information leaks in federated learning.
\newblock In {\em Proceedings of the Network and Distributed System Security
  Symposium}, volume~10, 2020.

\bibitem[\protect\citeauthoryear{Rebuffi and et al}{2017}]{rebuffi2017icarl}
Sylvestre-Alvise Rebuffi and et~al.
\newblock icarl: Incremental classifier and representation learning.
\newblock In {\em CVPR}, pages 2001--2010, 2017.

\bibitem[\protect\citeauthoryear{Rieke and et al}{2020}]{rieke2020future}
Nicola Rieke and et~al.
\newblock The future of digital health with federated learning.
\newblock {\em NPJ digital medicine}, 3(1):1--7, 2020.

\bibitem[\protect\citeauthoryear{Roberts and et al}{2020}]{roberts2020much}
Adam Roberts and et~al.
\newblock How much knowledge can you pack into the parameters of a language
  model?
\newblock {\em arXiv preprint arXiv:2002.08910}, 2020.

\bibitem[\protect\citeauthoryear{Samarakoon and et
  al}{2019}]{samarakoon2019distributed}
Sumudu Samarakoon and et~al.
\newblock Distributed federated learning for ultra-reliable low-latency
  vehicular communications.
\newblock {\em IEEE TC}, 68(2):1146--1159, 2019.

\bibitem[\protect\citeauthoryear{Schmidhuber}{2015}]{schmidhuber2015deep}
J{\"u}rgen Schmidhuber.
\newblock Deep learning in neural networks: An overview.
\newblock {\em Neural networks}, 61:85--117, 2015.

\bibitem[\protect\citeauthoryear{Shoham and et al}{2019}]{shoham2019overcoming}
Neta Shoham and et~al.
\newblock Overcoming forgetting in federated learning on non-iid data.
\newblock {\em arXiv:1910.07796}, 2019.

\bibitem[\protect\citeauthoryear{Shokri and et al}{2017}]{shokri2017membership}
Reza Shokri and et~al.
\newblock Membership inference attacks against machine learning models.
\newblock In {\em 2017 IEEE symposium on security and privacy (SP)}, pages
  3--18. IEEE, 2017.

\bibitem[\protect\citeauthoryear{Sun and et al}{2019}]{sun2019can}
Ziteng Sun and et~al.
\newblock Can you really backdoor federated learning?
\newblock {\em arXiv preprint arXiv:1911.07963}, 2019.

\bibitem[\protect\citeauthoryear{Thrun}{1995}]{thrun1995lifelong}
Sebastian Thrun.
\newblock A lifelong learning perspective for mobile robot control.
\newblock In {\em Intelligent robots and systems}, pages 201--214. Elsevier,
  1995.

\bibitem[\protect\citeauthoryear{Usmanova and et
  al}{2022}]{usmanova2022federated}
Anastasiia Usmanova and et~al.
\newblock Federated continual learning through distillation in pervasive
  computing.
\newblock In {\em IEEE SMARTCOMP}, pages 86--91, 2022.

\bibitem[\protect\citeauthoryear{Wang and et al}{2020}]{wang2020attack}
Hongyi Wang and et~al.
\newblock Attack of the tails: Yes, you really can backdoor federated learning.
\newblock {\em Advances in Neural Information Processing Systems},
  33:16070--16084, 2020.

\bibitem[\protect\citeauthoryear{Wang and et al}{2021}]{wang2021addressing}
Lixu Wang and et~al.
\newblock Addressing class imbalance in federated learning.
\newblock In {\em AAAI}, volume~35, pages 10165--10173, 2021.

\bibitem[\protect\citeauthoryear{Wang and et
  al}{2022}]{10.1145/3485447.3512222}
Junxiao Wang and et~al.
\newblock Federated unlearning via class-discriminative pruning.
\newblock In {\em Proceedings of the ACM Web Conference 2022}, pages 622--632,
  2022.

\bibitem[\protect\citeauthoryear{Warnat-Herresthal and et
  al}{2021}]{warnat2021swarm}
Stefanie Warnat-Herresthal and et~al.
\newblock Swarm learning for decentralized and confidential clinical machine
  learning.
\newblock {\em Nature}, 594(7862):265--270, 2021.

\bibitem[\protect\citeauthoryear{Wu and et al}{2019}]{wu2019large}
Yue Wu and et~al.
\newblock Large scale incremental learning.
\newblock In {\em CVPR}, pages 374--382, 2019.

\bibitem[\protect\citeauthoryear{Wu and et al}{2022a}]{kdunlearning}
Chen Wu and et~al.
\newblock Federated unlearning with knowledge distillation.
\newblock {\em arXiv preprint arXiv:2201.09441}, 2022.

\bibitem[\protect\citeauthoryear{Wu and et al}{2022b}]{leijie}
Leijie Wu and et~al.
\newblock Federated unlearning: Guarantee the right of clients to forget.
\newblock {\em IEEE Network}, 36(5):129--135, 2022.

\bibitem[\protect\citeauthoryear{Xu and et al}{2022}]{xu2022client}
Xinlei Xu and et~al.
\newblock Client selection based weighted federated few-shot learning.
\newblock {\em Applied Soft Computing}, 128:109488, 2022.

\bibitem[\protect\citeauthoryear{Yan and et al}{2022}]{ijcai2022p556}
Haonan Yan and et~al.
\newblock Arcane: An efficient architecture for exact machine unlearning.
\newblock In Lud~De Raedt, editor, {\em IJCAI}, pages 4006--4013, 7 2022.

\bibitem[\protect\citeauthoryear{Yang and et al}{2019}]{yang2019federated}
Qiang Yang and et~al.
\newblock Federated learning.
\newblock 13(3):1--207, 2019.

\bibitem[\protect\citeauthoryear{Yang and et al}{2021}]{yang2021flop}
Qian Yang and et~al.
\newblock Flop: Federated learning on medical datasets using partial networks.
\newblock In {\em ACM SIGKDD}, pages 3845--3853, 2021.

\bibitem[\protect\citeauthoryear{Yao and Sun}{2020}]{yao2020continual}
Xin Yao and Lifeng Sun.
\newblock Continual local training for better initialization of federated
  models.
\newblock In {\em ICIP}, pages 1736--1740. IEEE, 2020.

\bibitem[\protect\citeauthoryear{Yoon and et al}{2021}]{yoon2021federated}
Jaehong Yoon and et~al.
\newblock Federated continual learning with weighted inter-client transfer.
\newblock In {\em ICML}, pages 12073--12086. PMLR, 2021.

\bibitem[\protect\citeauthoryear{Yu and et al}{2022}]{yu2022legonet}
Sihao Yu and et~al.
\newblock Legonet: A fast and exact unlearning architecture.
\newblock {\em arXiv preprint arXiv:2210.16023}, 2022.

\bibitem[\protect\citeauthoryear{Yuan and et al}{2023}]{recommendation}
Wei Yuan and et~al.
\newblock Federated unlearning for on-device recommendation.
\newblock In {\em ACM WSDM}, WSDM '23. Association for Computing Machinery,
  2023.

\bibitem[\protect\citeauthoryear{Zhang and et al}{2020}]{9155414}
Xiaoli Zhang and et~al.
\newblock Enabling execution assurance of federated learning at untrusted
  participants.
\newblock In {\em IEEE INFOCOM}, pages 1877--1886, 2020.

\bibitem[\protect\citeauthoryear{Zhang and et al}{2022}]{zhang2022survey}
Rui Zhang and et~al.
\newblock A survey on gradient inversion: Attacks, defenses and future
  directions.
\newblock {\em arXiv preprint arXiv:2206.07284}, 2022.

\end{thebibliography}

\end{document}